\documentclass[letterpaper, 11pt, conference]{IEEEtran}

\IEEEoverridecommandlockouts

\usepackage{graphicx}
\usepackage{booktabs}
\usepackage{array}
\usepackage{url}
\usepackage{xcolor}
\usepackage[colorlinks=true, urlcolor=blue!60!black, linkcolor=black, citecolor=black]{hyperref}

\newcolumntype{L}[1]{>{\raggedright\arraybackslash}p{#1}}

\title{\LARGE \bf HATCH: Hardware-Attested Task Chains for Verifiable Manipulation\thanks{Accepted at the 5th Workshop on Open Robotics Software, IROS 2026, Pittsburgh, PA, USA.}}

\author{\IEEEauthorblockN{Shiza Charania}
\IEEEauthorblockA{\textit{Midas Robotics Inc.}, San Francisco, USA \\
\textit{University of Waterloo}, Waterloo, Canada \\
{scharania@uwaterloo.ca}}
}

\begin{document}

\maketitle
\thispagestyle{empty}
\pagestyle{empty}

\begin{abstract}
Capability claims in robot manipulation are unverifiable, and existing leaderboards are each closed, simulated, or gameable. We propose HATCH, a robot escape room: an openly specified, physically replicable, sensor-scored, and unrehearsable benchmark, released under an open license.
\end{abstract}

\section{Why Manipulation Has No Trusted Benchmark}

Capability claims in robot manipulation are, at present, unverifiable. Results arrive as curated demonstration videos, run on hardware no other lab owns, in environments no other lab can recreate, against self-defined success criteria. Vision and language compounded progress on shared benchmarks; ImageNet is the defining case. Manipulation never acquired one, in part because physical evaluation is expensive: hardware, resets between runs, and human labor.

The leaderboards that do exist achieve at most two of the three properties a trusted benchmark requires: openness, physical evaluation, and trustworthy results. \emph{RoboLab-120}~\cite{robolab120} evaluates entrants in-house, in simulation, which resists gaming but is neither open nor physical: the entrant pool is small, and as of mid-2026, the operating vendor's own model topped the leaderboard~\cite{paxton_leaderboards}. \emph{RoboArena}~\cite{roboarena} distributes real-world evaluation across volunteer institutions---an open and physical design---but its team recently disclosed that volunteer evaluators had not completed some assigned evaluations, pulling affected results and flagging low-sample entries~\cite{roboarena_disclosure}; the failure was in the trust layer, not the robots. \emph{MolmoSpaces}~\cite{molmospaces} is fully open but simulation-only, and simulated success does not certify physical capability~\cite{rlbench}. Nothing available today is simultaneously open, physical, and trustworthy~\cite{paxton_leaderboards, paxton_progress}, so we propose one that is: HATCH, a robot escape room.

\section{Benchmark Design}

\emph{Why an escape room.} An escape room is a useful evaluation container for three reasons. First, it is \emph{bounded and exhaustively specifiable}: a finite set of objects, fixtures, and sensors in a defined volume, documentable well enough for a second lab to rebuild. Second, tasks chain when assembled into a room: the sensor confirming one task's completion releases the key, code, or clue for the next, so a room is one long-horizon evaluation rather than a suite of isolated trials. A run that fails partway still yields a score, but the headline result is escape. Third, it is \emph{adversarial by construction}: evaluators design the room and entrants see it for the first time at evaluation, so there is no home-field advantage and nothing to overfit.

A room instance draws from a task pool of five skill categories crossed with three difficulty levels (Table~\ref{fig:matrix}).

\begin{table*}[t]
\caption{HATCH task matrix (v0.1), one task per cell; full specifications in the open repository.}
\label{fig:matrix}
\centering
\scriptsize
\begin{tabular}{@{}L{2.6cm} L{4.4cm} L{4.6cm} L{4.8cm}@{}}
\toprule
\textbf{Category} & \textbf{Easy} & \textbf{Medium} & \textbf{Hard} \\
\midrule
Spatial reasoning & Place 4 labeled objects on matching colored mats & Fit chunky wooden shapes into a shape-sorting board & Rearrange 4 objects onto colored circles per a reference-card diagram \\
Symbolic / diagrammatic & Arrange blocks per a symbol-to-color key; color sensors confirm & Move a checker along a colored-arrow sequence; pressure plate confirms final square & UV light reveals $n$ symbols; select only matching objects from $m>n$ \\
Multi-step chain & Note directs robot to red shelf's letter; open compartment; retrieve clue & Match 5 of 10 objects to pinned photos; arrange in board order & Marked fake book hides key $\to$ box $\to$ order 4 tokens by size $\to$ enter revealed code on tile rack \\
Physical / causal & Open sealed envelope, extract keycard, swipe reader & Lift jar lids, read numbered tokens, place on numbered mat & Swap mismatched pen caps so body/cap colors match; overhead camera confirms \\
Dexterous / force & Insert varied-size pegs flush into matching holes; seating sensor confirms & Pick up flat poker chips; deposit each through a narrow slot, counter confirms & Match 4 screws to subtly different-sized holes; insert and drive each flush \\
\bottomrule
\end{tabular}
\end{table*}

\emph{Instrumented success.} Outcomes are confirmed by sensors, not human judgment: color sensors at placement positions, a pressure plate on the checker's final square, seating sensors confirming peg insertion, and an overhead camera checking the pen-cap task. A success signal a machine emits is one a third party can audit.

\emph{Selection under distractors.} Tasks present $m$ candidates when only $n<m$ are correct, so a robot must choose, not sweep. The UV task reveals only a subset of the symbols and the photo-matching task pins 5 photographs against 10 objects. Grasping everything scores zero.

\emph{Embodiment-agnostic scoring.} Scoring depends only on task outcomes and elapsed time. Arms, humanoids, and mobile manipulators compete on one leaderboard; the benchmark measures what was accomplished, not what embodiment accomplished it.

\emph{A room instance.} A medium-level room draws one medium task per category and chains them: seating the last shape in the sorting board (spatial) releases the arrow-sequence card; the checker's final square (symbolic) trips the pressure plate that unlocks the photo-matching station (multi-step), whose completed board order opens the token jars (physical/causal); completing the numbered mat releases the poker chips (dexterous), and the final chip dropping through the slot ends the run. Each stage's sensor both scores the stage and gates the next. The run is scored as stages completed plus time, with placements randomized by a run-day seed (Sec.~III).

\section{Openness and Verification}

Everything required to replicate the benchmark is released in the public repository (\url{https://github.com/shizacharania/hatch-benchmark}) under an open license: task specifications, a room build guide with bill of materials, scoring rubrics, difficulty-level definitions, and a standard result-reporting format. Objects are commodity items and, where possible, drawn from the YCB set~\cite{ycb}, keeping rebuild cost low. The v0.1 pool holds one task per cell, enough to assemble one room per difficulty level, varied between runs by seeded placements. Growing each cell to several interchangeable tasks, so room composition is itself sampled at evaluation, is next on the roadmap, alongside automated scoring and audit tooling.

Openness alone does not produce trust; verification is a spectrum of design choices. At one end sit \emph{organized events} in the DARPA-challenge lineage~\cite{darpa}: high trust, low scale. At the other end sits \emph{honor-system self-reporting}, which demonstrably breaks (RoboArena~\cite{roboarena_disclosure}). We propose \emph{attested self-runs} as the default middle: entrants rebuild the room from the public guide and submit uncut video, complete sensor logs, and the run-day randomization seed governing object placement. Because success is instrumented, the sensor record carries most of the trust burden, not a reviewer's eye. Occasional organized events then serve as ground-truth calibration for the self-run pool.

\section{Open Questions}

Treating a physical benchmark as open infrastructure imports the software community's maintenance questions regarding versioning, continuous integration, and governance into a domain where the artifact is a room. How should a physical benchmark be versioned, and what is a breaking change when old scores cannot be re-run? Who governs the task pool and audits submissions? How does the pool grow without new tasks leaking to entrants before evaluation? We offer HATCH v0.1 as a concrete first answer to the question upstream of all of these: manipulation needs a benchmark that is open, physical, and trustworthy at once, and no existing leaderboard is all three.

\end{document}